\documentclass{article}

\usepackage{amsmath}
\usepackage{arxiv}
\usepackage{hyperref}
\usepackage{multirow}
\usepackage{enumitem}
\usepackage[utf8]{inputenc} 
\usepackage[T1]{fontenc}    
\usepackage{hyperref}       
\usepackage{url}            
\usepackage{booktabs}       
\usepackage{amsfonts}       
\usepackage{nicefrac}       
\usepackage{microtype}      
\usepackage{lipsum}
\usepackage{graphicx}
\usepackage{xcolor}
\usepackage[normalem]{ulem}

\graphicspath{ {./images/} }

\title{Monitoring Covid-19 on social media using a novel triage and diagnosis approach}
\author{
	Abul Hasan\\
	Department of Computer Science\\
	Birkbeck, University of London\\
	London WC1E 7HX UK\\
	\texttt{abulhasan@dcs.bbk.ac.uk} \\
	\And
	Mark Levene\\
	Department of Computer Science\\
	Birkbeck, University of London\\
	London WC1E 7HX UK\\
	\texttt{mlevene@dcs.bbk.ac.uk} \\
	\And
	David Weston\\
	Department of Computer Science\\
	Birkbeck, University of London\\
	London WC1E 7HX UK\\
	\texttt{dweston@dcs.bbk.ac.uk} \\
	\And
	Renate Fromson\\
	Barnet General Hospital\\
	Wellhouse Lane\\
	London EN5 3DJ UK \\
	\texttt{rfromson4@gmail.com} \\
	\And
	Nicolas Koslover\\
	Barnet General Hospital\\
	Wellhouse Lane\\
	London EN5 3DJ UK \\
	\texttt{nic.koslover@hotmail.co.uk} \\
	\And
	Tamara Levene\\
	Barnet General Hospital\\
	Wellhouse Lane\\
	London EN5 3DJ UK \\
	\texttt{tamaralevene@gmail.com} \\
}

\begin{document}
	\maketitle
	\begin{abstract}
		\textbf{Background:} The COVID-19 pandemic has created a pressing need for integrating information from disparate sources, in order to assist decision makers. Social media is important in this respect, however, to make sense of the textual information it provides and be able to automate the processing of large amounts of data, natural language processing methods are needed. Social media posts are often noisy, yet they may provide valuable insights regarding the severity and prevalence of the disease in the population. Here, we adopt a triage and diagnosis approach to the analysis of social media posts using machine learning techniques for the purpose of disease detection and surveillance. We thus obtain useful prevalence and incidence statistics to identify disease symptoms and their severities, motivated by public health concerns.
		 \\
		\textbf{Objective:}This study aims to develop an end-to-end natural language processing pipeline for triage and diagnosis of COVID-19 from patient-authored social media posts, in order to provide researchers and public health practitioners with additional information on the symptoms, severity and prevalence of the disease rather than to provide an actionable decision at the individual level.\\
		\textbf{Materials and Methods:} The text processing pipeline first extracts COVID-19 symptoms and related concepts such as severity, duration, negations, and body parts from patients' posts using conditional random fields. An unsupervised rule-based algorithm is then applied to establish relations between concepts in the next step of the pipeline. The extracted concepts and relations are subsequently used to construct two different vector representations of each post. These vectors are applied separately to build support vector machine learning models to triage patients into three categories and diagnose them for COVID-19. \\
		\textbf{Results:} We report that macro- and micro-averaged $F_1$ scores in the range of 71-96\% and 61-87\%, respectively, for the triage and diagnosis of COVID-19, when the models are trained on human labelled data. Our experimental results indicate that similar performance can be achieved when the models are trained using predicted labels from concept extraction and rule-based classifiers, thus yielding end-to-end machine learning. Also, we highlight important features uncovered by our diagnostic machine learning models and compare them with the most frequent symptoms revealed in another COVID-19 data set. In particular, we found that the most important features are not always the most frequent ones.\\
		\textbf{Conclusions:} Our preliminary results show that it is possible to automatically triage and diagnose patients for COVID-19 from social media natural language narratives, using a machine learning pipeline in order to provide information on the severity and prevalence of the disease for use within health surveillance systems.\\
		\textbf{Key words:}  COVID-19, Disease detection and surveillance, Medical social media, Natural Language processing, Conditional random fields, Support vector machines.
	\end{abstract}
	\section*{Introduction}
	
	\label{sec:01}
	\subsection*{Overview}
	
	During the ongoing coronavirus pandemic, hospitals have been continuously at risk of being overwhelmed by the number of people developing serious illness. People in the UK were advised to stay at home if they had coronavirus symptoms and to seek assistance through NHS helpline if they needed to \cite{UKGoV}. Consequently, there is an urgent need to develop novel practical approaches to assist medical staff. A variety of methods have been recently developed that involve natural language processing techniques;  The concerns of these methods range from the level of the individual, see for example \cite{obeid2020, schwab2020}, up to the population level \cite{sarker2020self,qin2020prediction}.

	\par 
	Herein, we take a diagnostic approach and propose an end-to-end {\em  Natural Language Processing (NLP)} pipeline to automatically triage and diagnose COVID-19 cases from patient-authored medical social media posts. The triage may inform decision-makers about the severity of COVID-19, and diagnosis could help in gauging the prevalence of infections in the population. Attempting a clinical diagnosis of influenza, or in our case a diagnosis of COVID-19, purely on the information provided in a social media post is unlikely to be sufficiently accurate to be actionable on an individual level, since the quality of this information will be typically noisy and incomplete. However, it is not necessary to have actionable diagnoses at the individual level in order to identify interesting patterns at the population level, which may be useful within public health surveillance systems. For example, text messages from the microblogging site Twitter was used to identify influenza outbreaks \cite{aramaki2011twitter}. In addition Twitter data in conjunction with a US Center for Disease Control and Prevention data set was used to predict the percentage of influenza-like illness in the US population  \cite{hu2018prediction}.

\par  One of our key concerns is in the production of a high-quality 
	human labelled data set on which to build our pipeline. In the following we give a brief overview of our pipeline and how we developed our data set. The first step in the pipeline is attained by developing an annotation application that detects and highlights COVID-19 related symptoms with their severity and duration in a social media post, henceforth collectively termed as {\em concepts}. During the second step relations between symptoms and other relevant concepts are also automatically identified  and annotated. For example, {\em breathing hurts} is a symptom which is related to a body part {\em upper chest area}.
	
	\par
\begin{figure*}[h]
		\centering
		\includegraphics{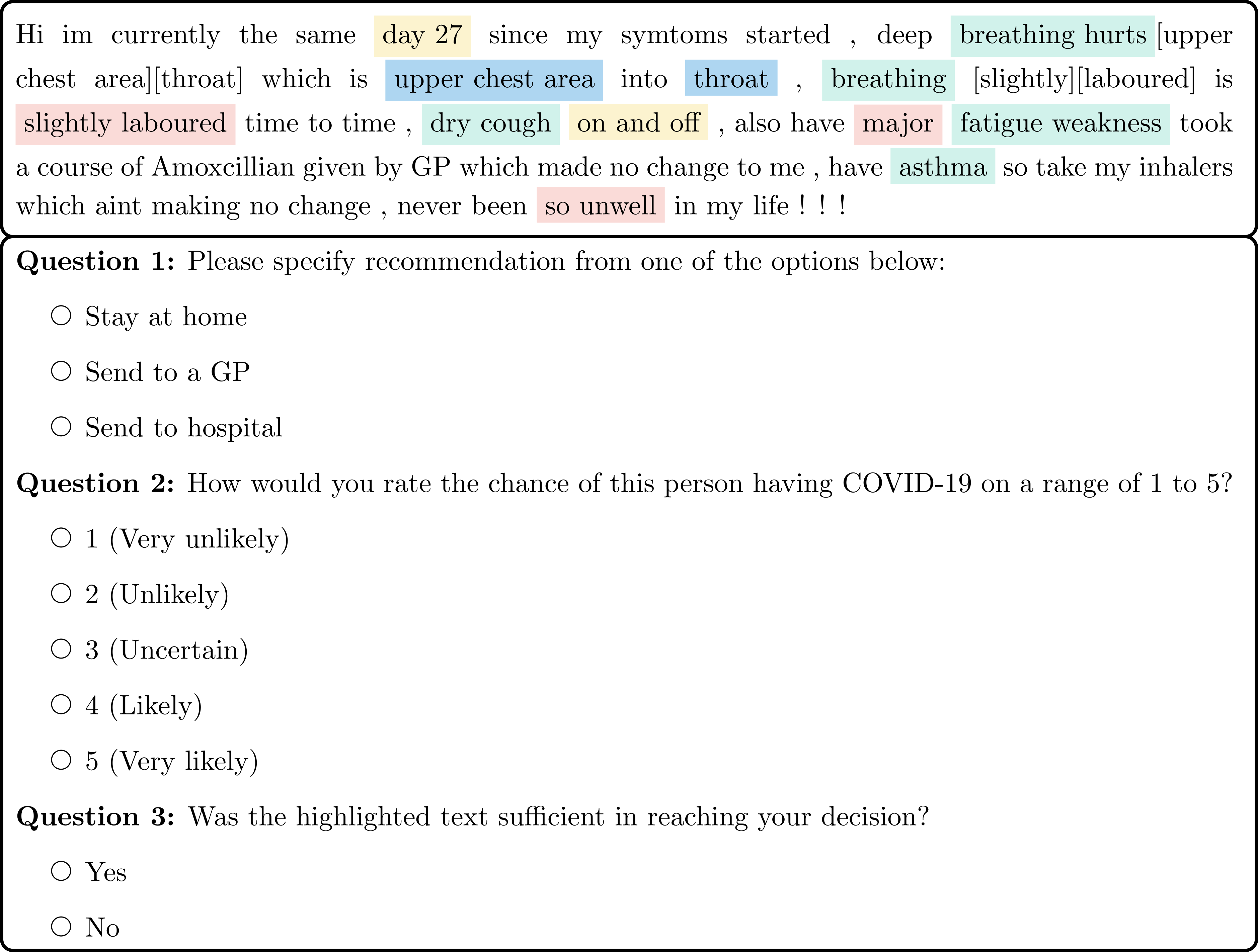}
		\caption{A patient-authored social media post is annotated with symptoms (light green), affected body parts (pale blue), duration (light yellow) and severities (pink). The phrases in the square brackets show relations between a symptom and a body part/duration/severity, when the distance was greater than 1. This annotated post was presented to three doctors to triage and diagnose the author of the post by answering {\em Questions 1} and {\em 2}, respectively.} \label{fig:01}
	\end{figure*}

	One author manually annotated our data with concepts and relations, allowing us to present posts highlighted with identified concepts and relations to three experts along with several questions, as shown in Figure ~\ref{fig:01}. The first question asked the experts to triage a patient into one of the following three categories: Stay at home, Send to a GP, and/or Send to hospital. The second question asked to diagnose the likelihood of COVID-19 in a Likert Scale of 1 to 5 \cite{norman2010likert}.
	\par The three experts are foundation doctors working in the UK who were redeployed to work on COVID-19 wards during the first wave of the pandemic, between March and July 2020. Their roles involved the diagnosis and management of patients with COVID-19, including patients who were particularly unwell and required either non-invasive or invasive ventilation. There were some training sessions organised for doctors working on COVID-19 wards. However, these were only provided towards the end of the first wave, as there was initially little knowledge of the virus and how to treat it. In the hospital the doctors followed local protocols, which were adjusted as more experience was gained about the virus. 
	
	\par 

We also asked the doctors to indicate whether the highlighted text presented is sufficient in reaching their decision, in order to understand its usefulness when we incorporate them in the annotation interface. The annotations were found to be sufficient in as many as 85\% of the posts, on average, as indicated by the doctors’ answers to Q3 in Figure ~\ref{fig:01}.
	
	\par The posts labelled by the doctors were then used to construct two types of predictive machine learning model using {\em Support Vector Machines} (SVM) \cite{drucker1996, marsland2015machine}; see Step 4 from section Methods.
	The {\em triage models} employ hierarchical binary classifiers, which consider the risk averseness or tolerance of the doctors when making the diagnosis \cite{arrieta2017risk}. 
	The {\em diagnostic models} first calculate the probability of a patient having COVID-19 from doctors' ratings. The probabilities are then used to construct three different decision functions for classifying {\em COVID} and {\em NO\_COVID} classes; these are detailed in the Problem setting subsection in Methods.
	
	\par We trained the SVM models in two different ways, first with ground truth annotations, and second using predictions from the concept and relation extraction step described above. Predictions  obtained from the concept extraction step make use of {\em Conditional Random Fields} (CRF) \cite{sutton}; see Step 1 of the Methodology Sub Section in Methods for implementation details. 
	Relations are obtained from these predicted concepts using an unsupervised {\em Rule-Based} (RB) classifier \cite{bach2007review}; see Step 2 from section Methods.
	
	
	\par We also discuss the feature importance obtained from the constructed COVID-19 diagnostic models, and compare them with the most frequent symptoms from \cite{sarker2020self} and our data set. We found that symptoms such as anosmia/ageusia (loss of taste and smell) rank in the top 5 most important features, whereas they do no rank in the top 5 most frequent symptoms; see Discussion. 
	Overall, we make several contributions as follows:
	\begin{enumerate}
		
		\item We show that it is possible to take an approach which aims at disease detection to augment public health surveillance systems, by constructing machine learning models to triage and diagnose COVID-19 from patients' natural language narratives. To the best of our knowledge, no other previous work has attempted to triage or diagnose COVID-19 from social media posts.
		\item We also build an end-to-end NLP pipeline by making use of automated concept and relation extraction. Our experiments show that the models built using predictions from concept and relation extraction produce similar results to those built using ground truth human concept annotation.
	\end{enumerate}
	
	\subsection*{Related work}
	
	\par 
	  Data derived from social media has been successfully used to facilitate the detection of influenza epidemics \cite{aramaki2011twitter, hu2018prediction}. In addition, \cite{edo2020scoping} provides a thorough review of the use of Twitter in public health surveillance for the purpose of monitoring, detecting and forecasting influenza-like illnesses. Since the start of the COVID-19 pandemic, a number of mobile app-based self-reported symptom tools have emerged, to track novel symptoms \cite{zens2020app}. The mobile application in \cite{menni2020real} applied Logistic Regression (LR)  to predict percentage of probable infected cases among the total app users in the US and UK combined. The authors in \cite{mizrahi2020longitudinal} performed a statistical analysis on primary care Electronic Health Records (EHR) data to find longitudinal dynamics of symptoms prior to and throughout the infection. 
		
	\par At an individual diagnostic level, Zimmerman et al. \cite{zimmerman2016} previously have applied Classification and Regression Trees to determine the likelihood of symptom severity of influenza in clinical settings. Moreover, machine learning algorithms, such as decision trees, have shown promising results in detecting COVID-19 from blood test analyses \cite{brinati2020detection}. Here, we focus on features extracted from a textual source to triage and diagnose COVID-19 for the purpose of providing population level statistics in the context of public health surveillance. Studies related to our work deploy features obtained from online portals, telehealth visits, and structured and unstructured patient/doctors notes from EHR. In general, COVID-19 clinical prediction models can broadly be categorised into risk, diagnosis and prognosis models \cite{wynants2020}.	
	\par In \cite{judson2020}, a portal-based COVID-19 self-triage and self-scheduling tool was employed to segment patients into four risk categories: emergent, urgent, no-urgent and self-care. Whereas, the online telemedicine system in \cite{liu2020covid} used LR  to  predict low, moderate and high risk patients, by utilising demographic information, clinical symptoms, blood tests and computed tomography (CT) scan results.
	
	\par In \cite{schwab2020}, various machine learning models were developed to predict patients outcome from clinical, laboratory and demographic features found in EHR \cite{kaggle}. They reported that Gradient Boosting (XGB), Random Forest and SVM are the best performing models for predicting COVID-19 test results, and, hospital and ICU admissions for positive patients, respectively. A detailed list of clinical and laboratory features can be found in \cite{wang2020clinical}, where they developed predictive models for the inpatient mortality in Wuhan, using an ensemble of XGB models. 
	Similarly, In \cite{vaid2020machine}, mortality and critical events for patients using XGB classifiers were predicted. Finally, a critical review on various diagnostic and prognostic models of COVID-19 used in clinical settings, can be found in \cite{wynants2020}. 
	
	\par In \cite{wagner2020augmented}, COVID-19 symptoms from unstructured clinical notes in the EHR of patients subjected to COVID-19 PCR testing were extracted. In addition, COVID-19 SignSym \cite{wang2020covid} was designed to automatically extract symptoms and related attributes from free text. Furthermore, the study in \cite{lopez2020covid} utilises radiological text reports form lung CT scans to diagnose COVID-19. Similar to our approach, Lopez et al. \cite{lopez2020covid} first extracted concepts using a popular medical ontology \cite{bodenreider2004} and then constructed a document representation using word embeddings \cite{mikolov2013} and concept vectors \cite{lopez2020covid}. However, our methodology differs from theirs with respect to the extraction of relations between concepts, and moreover, our data set, comprising posts obtained from medical social media, is more challenging to work with, since social media posts exhibit greater heterogeneity in language than radiological text reports.
	
	\par Finally, \cite{sarker2020self} published a COVID-19 symptom lexicon extracted from Twitter,  which we compare our work to in the Discussion section.

	\begin{table*}[h]\caption{Pair-wise agreement between pairs of doctors answers for Question 1 and 2; see Figure~\ref{fig:01} for an example. } \label{tab:01}
		
		\begin{center}
			\begin{tabular}[h!]{|c|c|c|c|c|c|c|c|} \cline{1-7} \hline
				\multicolumn{4}{|c|}{Question 1}&&\multicolumn{3}{c|}{Question 2} \\ \cline{1-4} \cline{6-8}
			Pair& {$p_o$}& Kappa& AC1&&{$p_o$}& Kappa& AC1\\ \cline{1-4} \cline{6-8} 
			AB&	0.65&	0.26&	0.55&&0.73&	0.64&	0.67 \\ \cline{1-4} \cline{6-8} 
			BC&	0.63&	0.14&	0.53&&0.73&	0.64&	0.67 \\ \cline{1-4} \cline{6-8} 
			AC&	0.77&	0.28&	0.72&&0.51&	0.40&	0.40 \\ \hline 
			\end{tabular}
		\end{center}
	\end{table*}
		
	\section*{Methods}
	\label{sec:02}
	\subsection*{Data}
	\label{subsec:02_1}
	\begin{figure}
		\centering
		\includegraphics[scale=0.5]{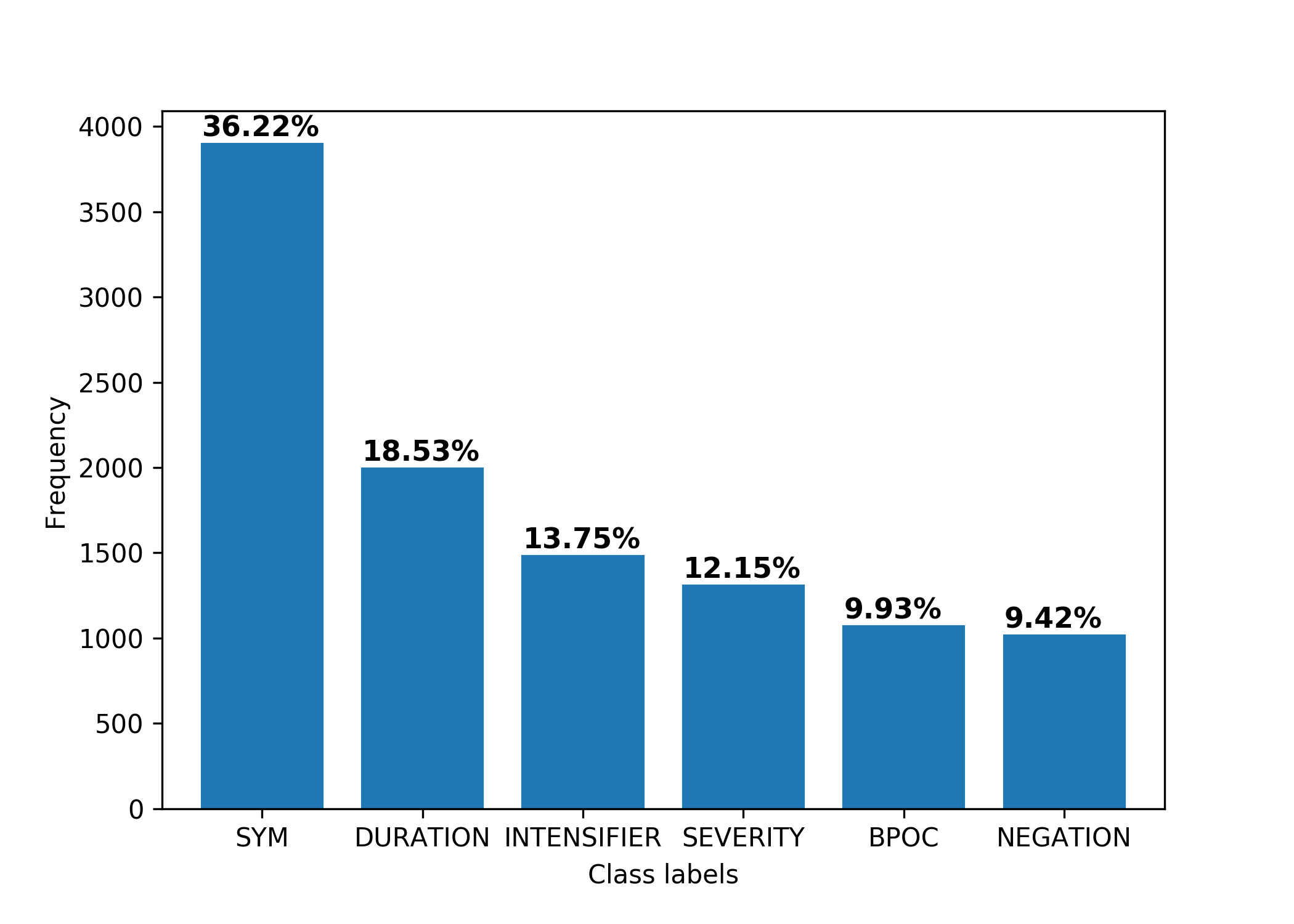}
		\caption{Frequency distribution of annotated classes/concepts from the text are shown. We also show the percentage of each class after discounting the {\em OTHER} labels. The average number of tokens per post is 130.17(SD = 97.83). Here, {\em SYM, DURATION, INTENSIFIER, SEVERITY, BPOC and NEGATION} denote symptoms, duration, intensifiers, severity, body parts and negations, respectively.
		} \label{fig:02}
	\end{figure}
	We collected social media posts discussing COVID-19 medical conditions from a forum called {\it Patient} \cite{PI}. This a public forum that was created at the onset of the coronavirus outbreak in the UK.
	 
	We obtained permission from the site administrator to scrape publicly available posts dated between April and June, 2020. In addition, all user IDs and metadata were removed from the posts for the purpose of the study. After the posts were anonymised, and the duplicates were removed, we randomly selected 500 distinct posts. The first author annotated these posts with the classes shown in Figure~\ref{fig:02}. The class labels represent symptoms and the related concepts: (i) duration, (ii) intensifier, which increases the level of symptom severity, (iii) severity, (iv) negation, which denotes presence or absence of the symptom or severity, and (v) affected body parts. We also annotated relations between a symptom and other concepts that exist at sentence level. For example, the relation between a symptom and a severity concept is denoted as {\em (SYM, SEVERITY)}. The posts were then marked with concepts in different colours and the relations were placed right after the symptom in square brackets, as shown in  Figure ~\ref{fig:01}. Each marked post was presented to the doctors using a web application, and they were asked three questions independently; see Figure~\ref{fig:01}. We call the doctors answers to Q1 and Q2 as the COVID-19 symptom triage and diagnosis, respectively. Thus for each post we have three independent answers from three doctors, which we denote as A, B, and C, respectively; these correspond to the last three authors of the paper and have been assigned randomly. 
	
	\subsubsection*{\textit{Measurement of agreement}}
	In order to measure the agreement between the answers (recommendations and ratings) of the three doctors to Q1 and Q2 of Figure ~\ref{fig:01}, we first calculated the proportion of observed agreement ($p_o$) as suggested by \cite{de2013clinicians}, who stipulate that Kappa is actually a measure of reliability rather than agreement, and observe that $p_o$ is high in all cases as can be seen in Table ~\ref{tab:01}. We note that paradoxical behaviour of Cohen's Kappa can arise when the absolute agreement ($p_o$) is high \cite{feinstein1990high}. This may occur when there is a substantial imbalance in the marginal totals of the answers, which we have observed in the answers to Q1. Consequently, we deploy a common solution to this problem called the AC1 statistic devised by Gwet \cite{gwet2008computing, wongpakaran2013comparison}, in addition to Cohen's Kappa.
	
	
	\par We found that for Q1 the AC1 measure shows moderate agreement (in the middle of the moderate range) between A and B (0.55), between B and C (0.53), and substantial agreement between A and C (0.72); see \cite{landis1977measurement} for benchmark scale for the strength of agreement. For Q2 it turns out that the said paradox did not occur, resulting in similar values for Kappa and AC1. The agreement between A and B (Kappa=0.64, AC1=0.67) and between B and C (Kappa=0.64, AC1=0.67) are substantial, while the agreement between A and C (Kappa=0.40, AC1=0.40) is on the boundary of fair and moderate; see Table ~\ref{tab:01}.

	\par It is important to note that COVID-19 is a novel virus, for which the doctors did not have prior experience or training before the first wave of the pandemic, and thus one would expect some difference of opinion. (We bear in mind that in our setting the doctors can only see the posts and thus cannot interact with the patients as they would in a normal scenario.) Moreover, there are probable differences in risk tolerances between the doctors, which would lead to potentially different decisions and diagnoses.

	\begin{figure*}[h]
		\includegraphics{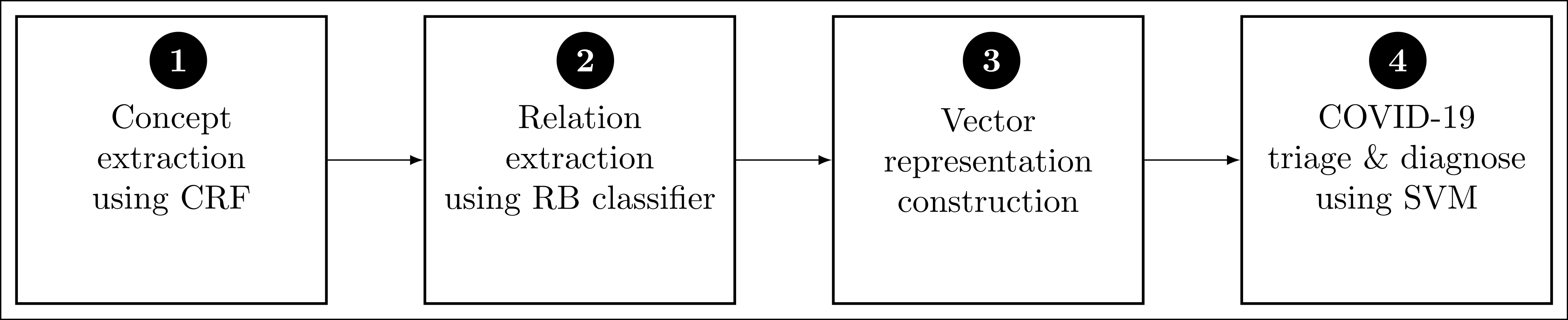}
		\caption{A block diagram of COVID-19 triage and diagnosis text processing pipeline. Here, CRF, RB classifier and SVM are acronyms for Conditional Random Fields, Rule-Based classifier and Support Vector Machine, respectively.} \label{fig:03}
	\end{figure*}
	
	\subsection*{Problem setting}
	\label{subsec:02_2}
	\subsubsection*{\textit{Triage classification for Question 1}}
	We map the doctors' recommendation from Q1 to ordinal values; the options {\it Stay at home}, {\it Send to a GP}, or {\it Send to hospital} are transformed to the values 1, 2, and 3, respectively. 
	In order to combine recommendations from two or more doctors, we first take their average. This result is rounded to an integer in one of two ways, either by taking the floor or the ceiling. Considering the risk attitude prevalent among medical practitioners \cite{arrieta2017risk}, we categorise the ceiling of the average to be {\it  risk averse},  denoted by e.g. AB(R-a), and the floor to be {\it risk tolerant}, denoted by e.g. AB(R-t). Thus for each patient’s post, we have in total eleven recommendations from three doctors for Q1, the full enumeration can be seen in the first column of Table ~\ref{tab:03}.
	We construct a hierarchical classification model for each of these recommendations, where the goal is to classify a post into one of the three options.
	
	\subsection*{\textit{Diagnosis classification for Question 2}}
	To diagnose whether a patient has COVID-19 from his or her post, we first estimate the probability of having the disease by normalising the rating, i.e given a rating, $r$, the probability of COVID-19, $Pr(COVID|r)$, which we term as the {\em ground truth probability} (abbreviated {\em GTP}),  is simply:
	\begin{equation*} \label{eq:01}
		Pr(COVID|r)=\frac{r-1}{4}.
	\end{equation*}
	\par 
	
	\par  Given our ground truth probability estimates are discrete we investigate three decision boundaries  based on a threshold value of 0.5 to classify a post as follows: 
	\begin{enumerate}[align=left]
		\item[$LE$:]  If $Pr(COVID|r)<=0.5$, then {\em NO\_COVID}, else {\em COVID}.
		\item[$LT$:] If $Pr(COVID|r)<0.5$, then {\em NO\_COVID}, else {\em COVID}.
		\item[$NEQ$:]  If $Pr(COVID|r)<0.5$ then {\em NO\_COVID},  \\  else if $Pr(COVID|r)>0.5$ then {\em COVID}. 
	\end{enumerate}
	Note  $NEQ$ differs to the other decisions in that we ignore those cases on the 0.5 boundary.

	
	\subsection*{Methodology}
	\label{subsec:02_3}
	
	A schematic of our methodology to triage and diagnose patients from their social posts is shown in Figure~\ref{fig:03}. Here, the circles denote the steps followed in the pipeline. We now detail each of these steps.
	
	\subsubsection*{\textit{Step 1: Concept extraction}}
	In the first step, we pre-process each patient's post by splitting it into sentences and tokens using the GATE software \cite{gate}  built-in {\em Natural Language Processing} (NLP) pipeline. For each token in a sentence we build discrete features that signal whether the token is a member of one of the following dictionaries: (i) Symptom, (ii) Severity, (iii) Duration, (iv) Intensifier, and (v) Negation. 
	The dictionaries were built by analysing the posts while annotating them. We also utilise the MetaMap system \cite{bodenreider2004}, assuming that it contains all the necessary technical terms, to map tokens to three useful semantic categories: {\em  Sign or Symptom;  Disease or Syndrome; Body Part, Organ, or Organ Component}. Due to the assumption regarding medical terms, the system does not expect any new additional terms, and thus we are justified in extracting concepts and relations in pre-processing steps. The pre-processed text is then used to build a concept extraction module to recognise the classes, shown in Figure ~\ref{fig:02}, by applying a CRF  \cite{sutton}. A detailed description of our CRF training methodology can be found in \cite{hasan2020learning}. The extracted concepts are then used for our next step to recognise the relations between concepts.
	\subsubsection*{\textit{Step 2: Relation extraction}}

	The semantic relation between a symptom and other concepts, which we formally termed as {\em modifiers}, is resolved using an unsupervised RB classifier algorithm. We first filter all symptom and modifier pairs from a sentence within a predefined distance and then select the closest modifier to a symptom to construct a relation. In total, we extracted five kinds of relations as follows: {\em (SYM, SEVERITY), (SYM, DURATION), (SYM, BPOC), (SYM, NEGATION)} and {\em (SYM, ?)}.
	\par The severity modifiers are mapped to a scale of 1-5. The semantic meaning of the scale is: {\em very mild,  mild, moderate, severe} and {\em very severe}, respectively. The duration modifiers are also mapped to real values in chunks of weeks. So, for example, {\em 10 days} is mapped to the value {\em 1.42}.
	
	\subsubsection*{\textit{Step 3: Vector representation}}
	Fixed length vector representations suitable as input for SVM classifiers are built as follows. 	
	\subsubsection*{\textit{Symptom-only vector representation}}	
	\par Let $\langle s_0, s_1 \ldots, s_n \rangle$ be a vector of symptoms constructed from the symptom vocabulary, for our data set the number of unique symptom words/phrases $n = 871$. To construct the vector representation for a post, we extract the concept, {\em SYM}, and the relation {\em (SYM, NEGATION)}, and set $s_i$ to 1, 0, or -1, according to whether the symptom is present, not present, or negated, respectively.
	
	
	\subsubsection*{\textit{Symptom-modifier relation vector representation}}
	
	The symptom-modifier relation vector is a much larger vector than the symptom-only and comprises three appended vectors containing: (i) the absence or presence of $110$ unique body parts, (ii) the absence or value of a symptom duration, and
	(iii) the absence, negation or value or a symptom severity.

	\subsubsection*{\textit{Step 4: Triage and diagnosis}}
	We utilise SVM classification and regression models to triage and diagnose patients' posts, respectively, from the vector representations described above. 
	For Q1, the recommendation from a doctor or combination of doctors is the class label of the post; see section Problem setting in Methods for a description. To build a binary classifier, we first combine the {\em Send to a GP} and {\em Send to hospital} recommendations to represent a single class, {\em Send}. The SVM is trained to distinguish between the {\em Stay at home} and the {\em Send} options; we call this {\em SVM Classifier 1}. Next, the posts labelled as {\em Stay at home} are discarded and {\em SVM Classifier 2} is built utilising the remaining posts to classify the {\em Send to GP} and {\em Send to hospital} recommendations.This results in a hierarchical classifier for COVID-19 triage.
	
	\par For diagnosing COVID-19 cases, we deploy a variant of SVM, called {\em Support Vector Regression} (SVR) \cite{drucker1996}, to estimate the probability of COVID-19. We use the  GTP that is derived from answers to Q2, as the dependent variable. SVR  takes as input a high dimensional feature vector such as a {\em symptom-only} or {\em symptom-modifier relation} vector representation, as described above. Classification is performed using the  three decision functions, $LE$, $LT$, and $NEQ$, described previously.
	
	\begin{table*}[ht]\caption{ The concept extraction using CRF and relation extraction using RB classifier results on 3-fold cross validation.} \label{tab:02}
		\begin{center}
			
			\begin{tabular}[h!]{|c|c|c|c|c|c|c|c|c|c|c|c|c|} \cline{1-5} \hline
				\multicolumn{5}{|c|}{Concept extraction using CRF}&&\multicolumn{7}{c|}{Relation extraction using RB classifier} \\  \cline{7-13}
				\multicolumn{5}{|c|}{}&&&\multicolumn{3}{c|}{With stop words}	&\multicolumn{3}{c|}{Without stop words} \\ \cline{1-5} \cline{7-13}
				Label &	 P 	& R &	 $F_1$& 	 Support&&Distance&P& R& $F_1$&P& R& $F_1$ \\ \cline{1-5} \cline{7-13}
				SYM &		 0.94 &		 0.97 &		 0.95 &		 1300&&2 & 0.74 & 0.63 & 0.68& 0.74 & 0.64 & 0.69\\  \cline{1-5} \cline{7-13}
				SEVERITY &		 0.80 &		 0.79 &		 0.79 &		 437&&3 & 0.75 & 0.67 & 0.71& 0.75 & 0.67 & 0.71\\  \cline{1-5} \cline{7-13}
				BPOC &		 0.92 &		 0.83 &		 0.87 &		 356&&4  & 0.75 & 0.69 & 0.72 & 0.75 & 0.69 & 0.72\\  \cline{1-5} \cline{7-13}
				DURATION &		 0.87 &		 0.91 &		 0.89 &		 667&&5 & 0.75 & 0.71 & 0.73& 0.74 & 0.71 & 0.73\\  \cline{1-5} \cline{7-13}
				INTENSIFIER &		 0.88 &		 0.97 &		 0.92 &		 494&&6 & 0.74 & 0.72 & 0.73 & 0.74 & 0.72 & 0.73\\  \cline{1-5} \cline{7-13}
				NEGATION &		 0.83 &		 0.89 &		 0.86 &		 338&&7 & 0.73 & 0.73 & 0.73 & 0.73 & 0.73 & 0.73\\  \cline{1-5} \cline{7-13}
				OTHER &		 0.99 &		 0.98 &		 0.98 &		 16892 &&\multicolumn{7}{c|}{}\\  \cline{1-5}
				Macro-average &		 0.89 &		 0.89 &		 0.89&&&\multicolumn{7}{c|}{}\\  \hline
			\end{tabular}
			
		\end{center}
	\end{table*}
		\begin{table*}[h]\caption{Question 1: Hierarchical classification results for RBF kernel using the symptom-modifier relation vector.} \label{tab:03}
		\begin{center}
			\begin{tabular}{|c|c|c|c|c|c|c|c|c|c|c|c|c|c|} \hline
				\multicolumn{14}{|c|}{Symptom-modifier relation vector} \\ \hline
				\multicolumn{7}{|c|}{Trained on the ground truth} &&\multicolumn{6}{c|}{Trained on the CRF predictions}\\ \cline{1-7} \cline{9-14}
				\multirow{2}{*}{Model}& \multicolumn{3}{c|}{SVM Classifier 1}& \multicolumn{3}{c|}{SVM Classifier 2}&& \multicolumn{3}{c|}{SVM Classifier 1}& \multicolumn{3}{c|}{SVM Classifier 2} \\ \cline{2-7} \cline{9-14}
				&P &     R &     $F_1$ &  P &     R &     $F_1$  && $P$ &     R &     $F_1$   &  P &     R &     $F_1$ \\ \cline{1-7} \cline{9-14}
				A&0.82&0.91&0.86&0.73&0.95&0.83&&0.81&0.89&0.85&0.72&0.91&0.80\\ \cline{1-7} \cline{9-14}
				B&0.73&0.77&0.75&0.81&0.99&0.89&&0.74&0.74&0.74&0.81&0.99&0.89\\ \cline{1-7} \cline{9-14}
				C&0.85&0.98&0.91&\textemdash&\textemdash&\textemdash&&0.85&0.96&0.90&\textemdash&\textemdash&\textemdash\\ \cline{1-7} \cline{9-14}
				AB(R-a)&0.70&0.75&0.72&0.80&0.96&0.88&&0.73&0.71&0.71&0.81&0.96&0.88\\ \cline{1-7} \cline{9-14}
				AB(R-t)&0.84&0.96&0.89&0.85&1.00&0.92&&0.84&0.94&0.88&0.84&1.00&0.92\\ \cline{1-7} \cline{9-14}
				BC(R-a)&0.72&0.75&0.73&0.92&1.00&0.96&&0.74&0.71&0.72&0.92&1.00&0.96\\ \cline{1-7} \cline{9-14}
				BC(R-t)&	0.86 &0.99&0.92&  \textemdash&\textemdash&\textemdash&&0.88&0.98&0.93&  \textemdash&\textemdash&\textemdash\\ \cline{1-7} \cline{9-14}
				AC(R-a)&0.79&0.87&0.83&0.89&1.00&0.94&&0.81&0.85&0.83&0.89&1.00&0.94\\ \cline{1-7} \cline{9-14}
				AC(R-t)&0.88&0.98&0.93&  \textemdash&\textemdash&\textemdash&&0.88&0.98&0.93& \textemdash&\textemdash&\textemdash\\ \cline{1-7} \cline{9-14}
				ABC(R-a)&0.70&0.76&0.73&0.89&0.99&0.93&&0.72&0.72&0.72&0.89&1.00&0.94\\ \cline{1-7} \cline{9-14}
				ABC(R-t)&0.88&0.99&0.93& \textemdash&\textemdash&\textemdash&&0.89&0.98&0.93& \textemdash&\textemdash&\textemdash\\ \hline
				\multicolumn{14}{|c|}{Symptom-only vector} \\ \hline
				\multicolumn{7}{|c|}{Trained on the ground truth} &&\multicolumn{6}{c|}{Trained on the CRF predictions}\\ \cline{1-7} \cline{9-14}
				\multirow{2}{*}{Model}& \multicolumn{3}{c|}{SVM Classifier 1}& \multicolumn{3}{c|}{SVM Classifier 2}&& \multicolumn{3}{c|}{SVM Classifier 1}& \multicolumn{3}{c|}{SVM Classifier 2} \\ \cline{2-7} \cline{9-14}
				&P &     R &     $F_1$ &     P &     R &     $F_1$&  &P &     R &     $F_1$ &     P &     R &     $F_1$\\ \cline{1-7} \cline{9-14}
				A&0.83&0.91&0.87&0.74&0.85&0.79&&0.84&0.89&0.87&0.74&0.82&0.78\\ \cline{1-7} \cline{9-14}
				B&0.71&0.81&0.76&0.81&0.98&0.89&&0.74&0.79&0.77&0.82&0.98&0.89\\ \cline{1-7} \cline{9-14}
				C&0.87&0.97&0.92&  \textemdash&\textemdash&\textemdash&&0.86&0.95&0.90&  \textemdash&\textemdash&\textemdash\\ \cline{1-7} \cline{9-14}
				AB(R-a)&0.69&0.75&0.72&0.83&0.96&0.89&&0.72&0.76&0.73&0.83&0.92&0.87\\ \cline{1-7} \cline{9-14}
				AB(R-t)&0.85&0.94&0.89&0.85&1.00&0.92&&0.87&0.93&0.90&0.84&0.98&0.90\\ \cline{1-7} \cline{9-14}
				BC(R-a)&0.71&0.79&0.75&0.92&0.99&0.95&&0.72&0.78&0.75&0.92&0.99&0.95\\ \cline{1-7} \cline{9-14}
				
				BC(R-t)&0.88&0.98&0.93&  \textemdash&\textemdash&\textemdash&&0.87&0.97&0.92& \textemdash&\textemdash&\textemdash\\ \cline{1-7} \cline{9-14}
				
				AC(R-a)&0.80&0.86&0.83&0.89&1.00&0.94&&0.80&0.86&0.83&0.89&1.00&0.94\\ \cline{1-7} \cline{9-14}
				
				AC(R-t)&0.90&0.98&0.94&   \textemdash&\textemdash&\textemdash&&0.89&0.95&0.92&  \textemdash&\textemdash&\textemdash\\ \cline{1-7} \cline{9-14}
				
				ABC(R-a)&0.68&0.74&0.71&0.90&1.00&0.95&&0.71&0.76&0.73&0.89&0.99&0.93\\ \cline{1-7} \cline{9-14}
				
				ABC(R-t)&0.90&0.98&0.94&    \textemdash&\textemdash&\textemdash&&0.90&0.95&0.92&  \textemdash&\textemdash&\textemdash\\ \cline{1-7} \hline
			\end{tabular}
		\end{center}
	\end{table*}
	\begin{table*}[h]\caption{ Question 2: Micro-averaged $F_1$ results for different models and decision functions. Here A, B, C are three medical doctors (abbreviated as Dr) who took part in the experiment.} \label{tab:04}
		\begin{center}
			\begin{tabular}[h!]{|c|c|c|c|c|c|c|c|c|c|c|c|c|c|}\hline
				\multicolumn{7}{|c|}{Trained on the ground truth} &&\multicolumn{6}{c|}{Trained on the CRF predictions}\\ \cline{1-7} \cline{9-14}
				\multirow{2}{*}{Model}&\multicolumn{3}{c|}{Symptom-modifier}& \multicolumn{3}{c|}{Symptom-only}&&\multicolumn{3}{c|}{Symptom-modifier}& \multicolumn{3}{c|}{Symptom-only}\\ \cline{2-7} \cline{9-14}
				&$LE$&$LT$&$NEQ$&$LE$&$LT$&$NEQ$&&$LE$&$LT$&$NEQ$&$LE$&$LT$&$NEQ$ \\ \cline{1-7} \cline{9-14}
				A &0.72&0.61&0.78&0.70&0.59&0.74&&0.68&0.64&0.76&0.50&0.79&0.74\\ \cline{1-7} \cline{9-14}
				B &0.78&0.61&0.76&0.78&0.62&0.77&&0.76&0.64&0.77&0.78&0.57&0.74\\ \cline{1-7} \cline{9-14}
				C &0.87&0.75&0.87&0.88&0.75&0.87&&0.86&0.75&0.87&0.87&0.74&0.86\\ \cline{1-7} \cline{9-14}
				AB &0.72&0.66&0.74&0.74&0.65&0.75&&0.70&0.65&0.73&0.71&0.66&0.74\\ \cline{1-7} \cline{9-14}
				BC &0.84&0.76&0.84&0.85&0.79&0.86&&0.83&0.76&0.83&0.85&0.78&0.86\\ \cline{1-7} \cline{9-14}
				AC &0.81&0.73&0.81&0.83&0.74&0.83&&0.80&0.74&0.82&0.80&0.73&0.81\\ \cline{1-7} \cline{9-14}
				ABC &0.74&0.67&0.76&0.75&0.67&0.77&&0.72&0.69&0.76&0.74&0.69&0.77\\ \hline
				
			\end{tabular}
		\end{center}
	\end{table*}
	\section*{Results}
	\label{sec:03}
	\subsection*{Evaluation}
	We evaluate the performance of the CRF and SVM classification algorithms using the standard measures of precision ($P$), recall ($R$) and
	macro- and micro-averaged $F_1$ scores \cite{manning1999foundations}. macro-averaged scores are computed by considering the score independently for each class and then taking the average, while micro-averaged scores are computed by considering all the classes together. 
	
\begin{figure*}[htb!]
		\centering
		\includegraphics[scale=0.50]{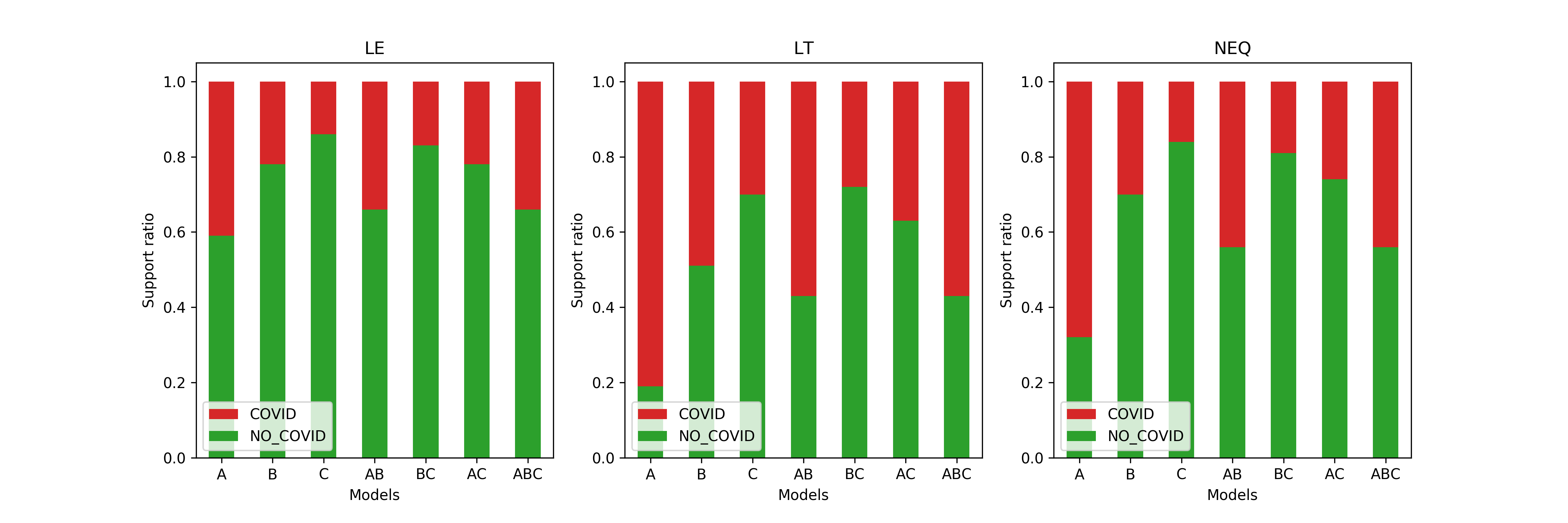}
		\caption{Support ratio of diagnosis classes across models and three decision functions for Question 2 classification tasks.} \label{fig:05}
	\end{figure*}

	As our data set is sufficiently balanced with {\em COVID} and {\em NO\_COVID} classes as can be seen in Figure ~\ref{fig:05}, we report micro-averaged scores for SVR classification. On the other hand, in case of concept extraction, the {\em Other} class dominates. So, we report the macro-averaged scores for the CRF classification results.
	
	
	\subsection*{Experimental setup}
	\label{subsec:03_1}
	
	For the CRF we report 3-fold cross validated macro-averaged results. Specifically, we trained each fold by a Python wrapper \cite{pycrf} for CRFsuite, see \cite{CRFsuite}. For relation extraction, we ran our unsupervised rule-based algorithm on the $500$ posts and calculated the $F_1$ scores by varying distances considering the two cases with and without stop words.
	
	
	\par We constructed SVM binary classifiers, {\em SVM Classifier 1} and {\em SVM Classifier 2}, using the Python wrapper for LIBSVM \cite{chang2011libsvm} implemented in Sklearn \cite{scikit-learn} with both Linear and Gaussian {\em Radial Basis Function} (RBF) kernels \cite{marsland2015machine}. 
	Similarly, the SVR \cite{SVR}, implemented using LIBSVM, is built with both Linear and RBF kernels. The hyperparameters ($C=10$ for the penalty,  $\gamma=0.01$ for the RBF kernel, and $\epsilon=0.5$ for the threshold) were discovered using grid search \cite{scikit-learn}. 
	
	
	\par We simulated two cases for COVID-19 triage and diagnosis. First SVM and SVR models trained with the ground truth examine the predictive performance when they are deployed as stand-alone applications. Second, when trained with the predictions from CRF and RB classifier, they resemble an end-to-end NLP application. To get a comparable result, the models were always tested with the ground truth. As a measure of performance, we report macro and micro-averaged $F_1$  scores for SVM classifiers and SVR, respectively.
	
	\begin{figure*}[htb!]
		\centering
		\includegraphics[scale=0.75]{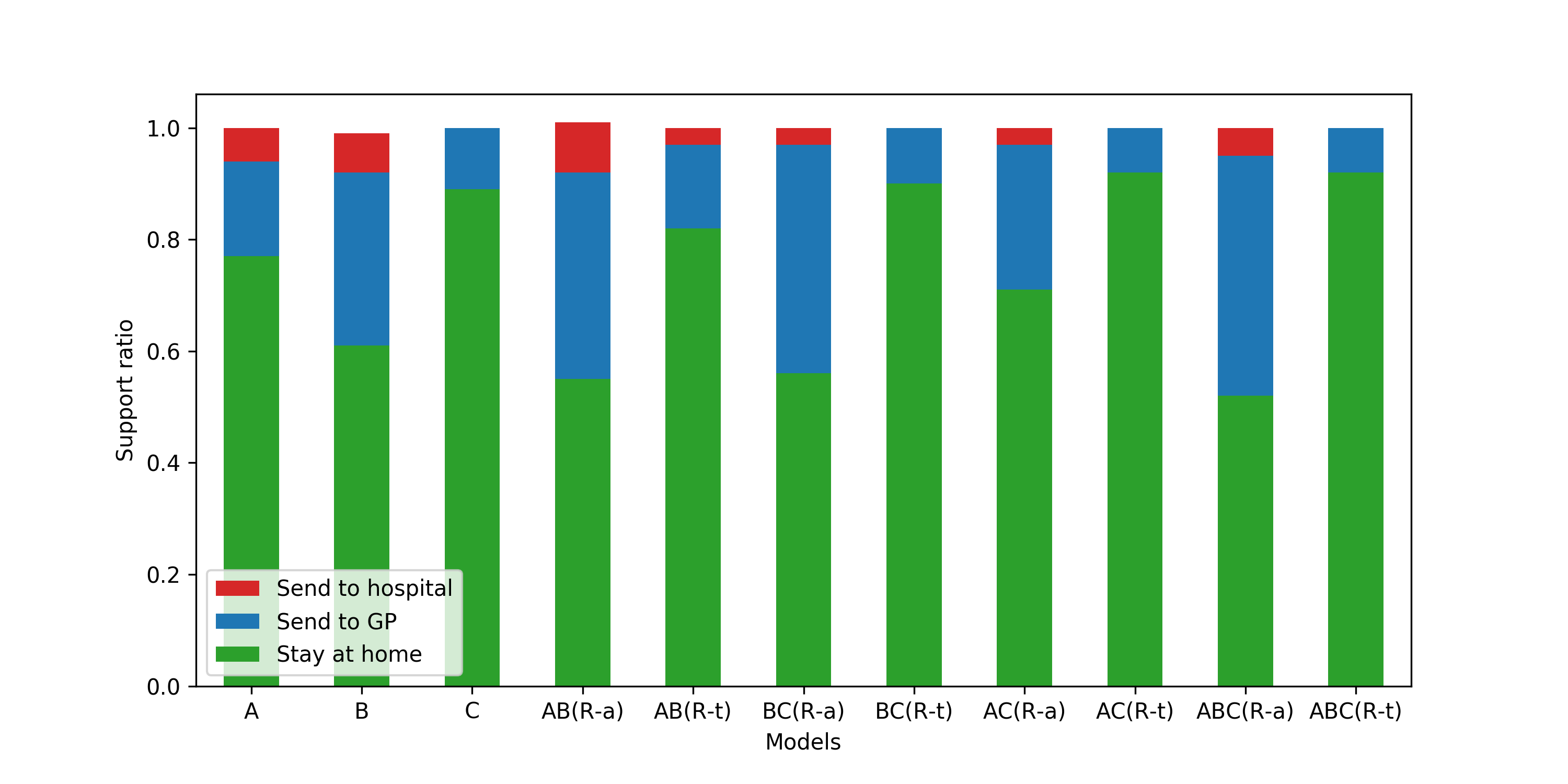}
		\caption{Support ratio of triage classes across models for Question 1 classification tasks. Absolute numbers for the {\em Send to hospital} class in test sets are as follows: A=10, B=12, AB(R-a)=14, AB(R-t)=5, BC(R-a)=6, CA(R-a)=5,  ABC(R-a)=9; the value for the remaining models is zero.} \label{fig:04}
	\end{figure*}
	
	\subsection*{Evaluation outcomes}
	\label{subsec:03_2}
	
	The concept and relation extraction phases produce excellent and very good predictive performances, respectively; see Table~\ref{tab:02}. The triage classification results from  Q1 are shown in Table~\ref{tab:03}. When we trained the models with the {\em Symptom-modifier vector} representations from the ground truth, the results of SVM Classifier 1 and 2 are in the range of 72-93\% and 83-96\%, respectively. The {Symptom-only vector} representations produces results in the range of 71-94\% and 79-95\%. These results suggest that we can achieve very good predictive performance for classifying {\em Stay at home} and {\em Send}, and for {\em Send to a GP} and {\em Send to hospital}. In general, risk-tolerant models achieve better performance than the risk-averse models. 
	However, since, in the test set, posts with the label Send to hospital are missing for some models (as can be seen from Figure ~\ref{fig:04}) we cannot report them. We report macro-averaged $F_1$ results since Q1 is framed as a decision problem, where weights for the classes are a priori equal. The results obtained after training with CRF predictions are in similar ranges for both representations and classifiers. This is important, because it indicates that an end-to-end NLP application is likely to produce similar predictive performance.
	
	\par Regarding Q2, when we trained the models with the {\em Symptom-modifier vector} representation from ground truth, the results of COVID-19 diagnosis are in the range of 72-87\%, 61-76\%, and 74-87\% for the $LE$, $LT$, and $NEQ$ decision functions, respectively; see Table~\ref{tab:04}. The {Symptom-only vector} representation produce results in the range of 70-88\%, 59-79\%, and 74-87\%. In general, $NEQ$ models perform better due to the omission of borderline cases where the GTPs are exactly 0.5. The support ratios for each model for different decision functions, is shown in Figure~\ref{fig:05}. When we trained the models with the {\em Symptom-modifier vector} representation from the CRF predictions, the results are in the range of 68-86\%, 64-76\%, and 73-87\% for the $LE$, $LT$, and $NEQ$ decision functions, respectively. This indicates that, for diagnosis as well as triage, an end-to-end NLP application is likely to perform similarly to standalone applications. Here, we report micro-averaged $F_1$ scores since, in our data set, {\em NO\_COVID} cases dominate; this largely resembles the natural distribution in the population, where people tested positive for coronavirus are relatively a low percentage in the whole population even when the prevalence of the virus is high. 
	
	\par Finally, we trained our models using a Linear kernel, but found that RBF dominates in most of the cases; however, Linear kernels are useful in finding feature importance \cite{weston2000feature}. 
		\begin{figure*}[htb!]
		\centering
		\includegraphics[scale=0.50]{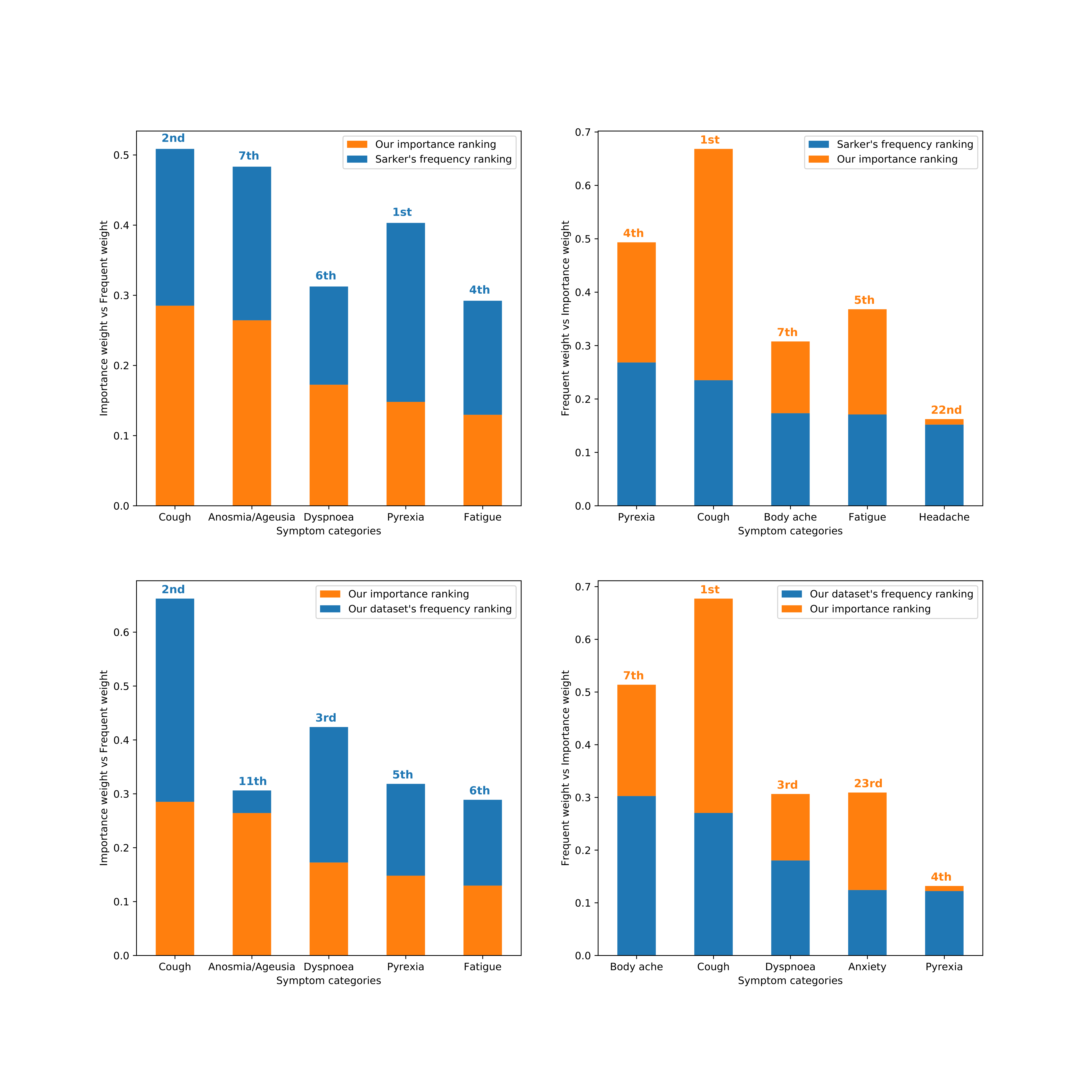}
		\caption{Feature comparison between our most important features and Sarker’s most frequent symptoms (top row), and between our most important features and our most frequent symptoms (bottom row). The feature importance rankings are obtained from an SVM linear kernel using the symptom-only vector representation.
		} \label{fig:06}
	\end{figure*}
	\section*{Discussion}
	\label{sec:04}
	\subsection*{Comparison with prior work}
		
	To quantify the important predictive features in the training set, we experimented with COVID-19 diagnosis using Linear Kernel SVR regression. More specifically, we used the Symptom-only vector representation constructed from the ground truth. We summed feature weights for each $s_i$ in  $\langle s_0, s_1 \ldots, s_n \rangle$ from seven models and three decision function; see Methods. The features are then mapped to the categories found in the Twitter COVID-19 lexicon complied by \cite{sarker2020self}. The top 5 important features in our data set are: {\em Cough, Anosmia/Agusia, Dyspnoea, Pyrexia and Fatigue}. In \cite{mizrahi2020longitudinal}, quoted 4 of these symptoms as the most prevalent coronavirus symptoms, strongly correlating with our findings.
	
	\par To compare our importance ranking with that of Sarker et al.'s \cite{sarker2020self} frequent categories, we compiled the corresponding frequencies of our 5 most important symptoms. 
	Normalised weights and frequencies are then plotted in  Figure \ref{fig:06}. The top-left stacked bar chart compares our 5 most important features with Sarker's frequencies. {\em Cough} is the most important symptom from our data set, where it is the $2^{nd}$ most frequent. {\em Anosmia/Aguesia} ranks $2^{nd}$ in our importance list, while it is $7^{th}$ in the most frequent list. {\em Pyrexia} comes $1^{st}$ and $4^{th}$ in both the frequent and importance lists, respectively.
	
	\par The top-right chart in Figure~\ref{fig:06} shows a comparison between Sarker's frequent ranking and our importance ranking. Here, we select top 5 most frequent symptoms from Sarker's frequency list and normalise them. They are: {\em Pyrexia, Cough, Body ache, Fatigue, and Headache}. We took the corresponding importance weights of those symptoms and plotted the stacked bar chart. Here, {\em Headache} ranks $22^{nd}$ in our importance ranking, while it is $5^{th}$ in the frequent ranking. We find a large difference between the two rankings, implying that the top most frequent symptoms are not necessarily the most important ones.
	
	\par Next we compare our most important feature weights with our data set's frequency ranking using the methods described above. From the bottom-left stacked bar chart of Figure ~\ref{fig:06}, we observe that Anosmia/Aguesia are a relatively low in order in the frequency ranking, i.e. $11^{th}$. Like  Sarker's, {\em Cough} comes $2^{nd}$ in our data set's frequency ranking.
	
	\par Finally, the bottom-right chart in Figure ~\ref{fig:06} refers to the comparison between our data set's frequency and importance rankings of the corresponding symptoms. We observe that {\em Anxiety} ranks $4^{th}$ in the most frequent list, where it is very low, i.e. $23^{rd}$, in the most importance ranking.
	\subsection*{Principal findings}
	This study demonstrates the potential to triage and diagnose COVID-19 patients from their social media posts. We have presented a proof of concept system to predict a patient’s health state by building machine learning models from their narrative. The models are trained in two ways; using (i) ground truth labels, and (ii) predictions obtained from the NLP pipeline. Trained models are always tested on the ground truth labels. We obtained good performances in both cases which indicates that an automated NLP pipeline could be used to triage and diagnose patients from their narrative; see Evaluation outcomes in the Results section. In general, health professionals and researchers could deploy, triage models to determine the severity of COVID-19 cases in the population, and diagnostic models to gauge the prevalence of the pandemic.  
	\subsection*{Limitations}
It is worth reiterating that social media posts, which are known to be noisy, are not on a par with the consultation that a patient would have with a doctor.  We stress that the aim of this study is to extract useful information at a population level, rather than to provide an actionable decision for an individual via social media posts. Our manually annotated data set has two main limitations. First having only three experts limits the quality of our labelling, although we deem this study to be a proof of concept. A larger number of experts, including more senior doctors would be beneficial in a follow-up study. The robustness of our results could be further improved by both increasing the size of our data set and introducing posts from several alternate sources. Given that the posts come from social media, it is not clear whether the results could be used as such in a diagnostic system, without combining them with actual consultations. However, it is worth noting that medical social media such as the posts we used herein, may uncover novel information regarding COVID-19.
	\subsection*{Conclusion}
	\label{sec:05}
The coronavirus pandemic has drawn a spotlight on the need to develop automated processes to provide additional information to researchers, health professionals and decision-makers. Medical social media comprises a rich resource of timely information that could fit this purpose. We have demonstrated that it is possible to take an approach which aims at the detection of COVID-19 using an automated triage and diagnosis system, in order to  augment public health surveillance systems, despite the heterogeneous nature of typical social media posts. The outputs from such an approach could be used to indicate the severity and estimate the prevalence of the disease in the population.

	\subsection*{Author Contribution}
	All authors were involved in the design of the work. The first author wrote the code. The first three authors drafted the article and all authors critically revised the article.
	\subsection*{Abbreviations}
	\begin{description}
		\item CRF: conditional random fields
		\item CT: computed tomography
		\item EHR: electronic health records
		\item LR: logistic regression
		\item NLP: natural language pipeline
		\item RBF: radial basis function
		\item SVM: support vector machine
		\item SVR: support vector regressions
		\item XGB: gradient boosting
	\end{description}

	\bibliographystyle{unsrt}  


\end{document}